\documentclass{article}
\usepackage{ijcai25}

\usepackage{times}
\usepackage{soul}
\usepackage{url}
\usepackage[hidelinks]{hyperref}
\usepackage[utf8]{inputenc}
\usepackage[small]{caption}
\usepackage{graphicx}
\usepackage{amsmath}
\usepackage{amsthm}
\usepackage{booktabs}
\usepackage{algorithm}
\usepackage{algorithmic}
\usepackage[switch]{lineno}
\usepackage{multirow}

\usepackage{amsfonts}
\usepackage{amssymb}

\title{ProtFlow: Fast Protein Sequence Design via Flow Matching on Compressed Protein Language Model Embeddings}

\author{
    Zitai Kong, Yiheng Zhu, Yinlong Xu, Hanjing Zhou, Mingzhe Yin, \\
    Jialu Wu, Hongxia Xu, Chang-Yu Hsieh, Tingjun Hou, Jian Wu
    \affiliations
    Zhejiang University
    \emails
    \{kongzitai, zhuyiheng2020, 22321336, jialuwu, Einstein, kimhsieh, tingjunhou, wujian2000\}@zju.edu.cn, \{zhj85393, mzyin256\}@gmail.com
}

\begin{document}

\maketitle

\begin{abstract}
    The design of protein sequences with desired functionalities is a fundamental task in protein engineering. Deep generative methods, such as autoregressive models and diffusion models, have greatly accelerated the discovery of novel protein sequences. However, these methods mainly focus on local or shallow residual semantics and suffer from low inference efficiency, large modeling space and high training cost. To address these challenges, we introduce \textsc{ProtFlow}, a fast flow matching-based protein sequence design framework that operates on embeddings derived from semantically meaningful latent space of protein language models. By compressing and smoothing the latent space, ProtFlow enhances performance while training on limited computational resources. Leveraging reflow techniques, ProtFlow enables high-quality single-step sequence generation. Additionally, we develop a joint design pipeline for the design scene of multichain proteins. We evaluate ProtFlow across diverse protein design tasks, including general peptides and long-chain proteins, antimicrobial peptides, and antibodies. Experimental results demonstrate that ProtFlow outperforms task-specific methods in these applications, underscoring its potential and broad applicability in computational protein sequence design and analysis.
\end{abstract}

\section{Introduction}
\label{sec:intro}

\begin{figure*}[t]
\centering
\includegraphics[width=0.8\textwidth]{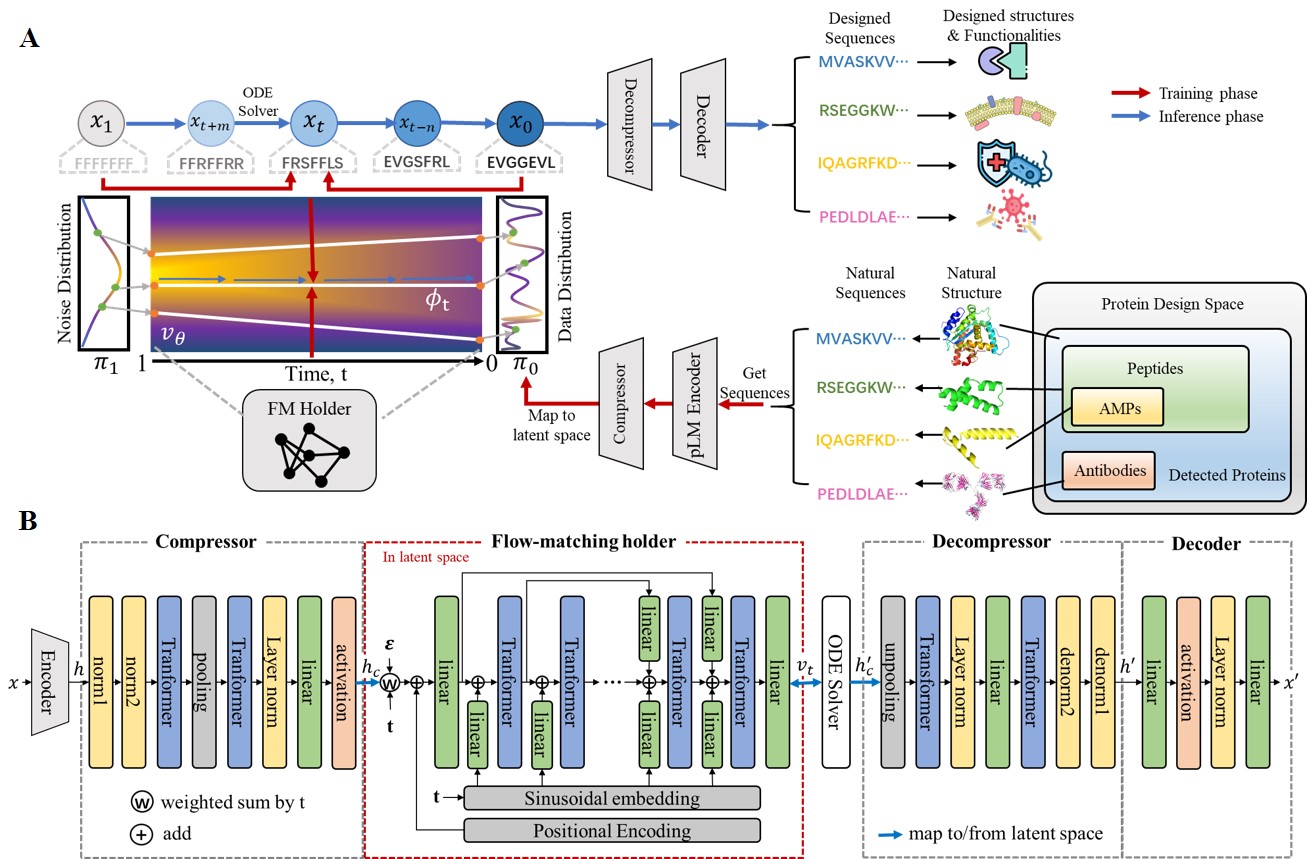} 
\caption{Overview of ProtFlow for protein sequence design. (A) Left: the visualization of the mathematical working flow of ProtFlow, including the training and inference phases. Right: relationships of different protein groups; (B) the architecture schematic of each components of ProtFlow. The FM holder is trained with other components frozen. In the training phase, the sampled sequence $x$ is mapped to the latent space as $h_c$ by the pLM encoder and compressor; the FM holder learns the FM vector field $v_t$. In the inference phase, the ODE solver starts from a sampled random noise $\epsilon$ and starting time point $t$, iteratively solves the $h'_c$ with $v_t$ represented by the FM holder, and maps back to $x'$ by the decompressor and decoder.}
\label{fig1}
\end{figure*}

Proteins are fundamental biomolecules that play crucial roles in organisms, serving as structural components of cells and facilitating various essential biological processes. The design of artificial proteins with desired functionalities has become a cornerstone of modern bioengineering research~\cite{huang2016coming}. However, the vast exploration space of possible protein sequences presents a significant challenge, and traditional biochemical approaches for discovering novel proteins are both time-intensive and expensive. Recently, the deployment of deep generative models has revolutionized protein design by enabling intelligent exploration of the biochemical landscape, offering a more efficient and cost-effective alternative~\cite{zhu2024generative}.

Given the similarities between protein sequences and natural language, advanced natural language processing (NLP) techniques, particularly autoregressive (AR) models, have been adapted for protein sequence design~\cite{ruffolo2024designing}. However, due to fundamental differences between protein sequences and natural language expressions, the unidirectional generation nature of AR methods may not always be optimal for protein design. Since non-autoregressive (NAR) approaches, predominantly diffusion models~\cite{ho2020denoising}, generate the entire sequence rather than predicting the next element step by step, they are better suited for modeling the long-range dependencies and complex amino acid interactions in proteins~\cite{wang2024diffusion}. Despite their advantages, diffusion models face several challenges, including prolonged generation times, high computational costs, and sensitivity to hyperparameters~\cite{dao2023flow,hu2024flow}. Although score-based diffusion methods have enhanced training efficiency, their reliance on a limited set of sampling probability paths necessitates specialized techniques, which still struggle to completely avoid sub-optimal outcomes~\cite{eijkelboom2024variational}.


Flow Matching (FM)~\cite{lipman2022flow} provides a more general and deterministic framework that directly learns the vector field guiding the optimal probability path from the prior noise distribution to the target data distribution, all in a simulation-free manner. FM achieves high-quality generation more efficiently by requiring significantly fewer ODE-solving steps than diffusion models. In spite of this efficiency, the optimal single-step generation can be further achieved by the Reflow technique~\cite{liu2022flow}. While FM has been successfully applied in various domains, including protein structure prediction~\cite{yim2024improved}, its potential for protein sequence design remains largely unexplored. A major challenge lies in adapting continuous FM models to accommodate the inherently discrete nature of protein sequences.


Inspired by latent diffusion models~\cite{rombach2022high}, we propose embedding discrete protein sequences into a latent space optimized for FM. Drawing inspiration from the transformative impact of large-scale language models in NLP~\cite{brown2020language}, Protein Language Models (pLMs), trained on evolutionary-scale protein sequences, have demonstrated significant potential for protein design~\cite{lin2022language}. Rather than developing an encoder from scratch, we leverage pLMs as encoders to operate within their semantically meaningful latent spaces, thereby enhancing both controllability and flexibility~\cite{hu2024flow}. However, pLM embeddings often suffer from excessively high dimensionality and massive activation issues~\cite{valeriani2024geometry}. To address these challenges, we redesign the pLM latent space by introducing a compressor that produces a more compact and smoothed latent representation.

Building on these insights, we introduce \textbf{ProtFlow}, a fast flow matching-based model for \textit{de novo} protein sequence generation on the compressed and smoothed pLM embedding space, which even supports 1-step generation with reflow. We also comprehensively evaluate the performance of the model in various practical protein design scenarios. We highlight our main contributions as follows:
\begin{itemize}
\item We introduce ProtFlow, the first flow matching-based generative model for protein sequence design to the best of our knowledge, which supports fast designs. 
\item We redesign the latent space of pLMs, achieving a 16-fold embedding compression, and experimentally demonstrate the improvements resulting from this optimized latent space.
\item We develop a multichain joint design pipeline and showcase ProtFlow's capability to generate structurally plausible, reliable, and natural protein sequences across various design tasks, including general peptides, long-chain proteins, antimicrobial peptides (AMPs), and antibodies, consistently outperforming existing methods.
\end{itemize}

\section{Related Work}
\label{sec:related_work}

\textbf{\textit{De novo} protein sequence design} focuses on constructing protein sequences that belong to biologically active groups, such as enzymes~\cite{yeh2023novo}, antibodies~\cite{frey2023protein,mahajan2023exploiting}, and peptides~\cite{chen2024amp} from scratch. Deep generative models have proven effective in designing high-quality \textit{de novo} protein sequences. Prominent methods include protein language models~\cite{lin2022language} and discrete diffusion models~\cite{alamdari2023protein,zhu2024bridgeif}. By encoding proteins into a latent space, advanced continuous diffusion models can be effectively applied to this task~\cite{meshchaninov2024diffusion}. Our work leverages this continuous approach to delve into the semantically rich latent protein space.

\textbf{Diffusion models}~\cite{ho2020denoising} can generate meaningful output by gradually denoising samples from prior noise distributions. To improve efficiency, the score-matching paradigm~\cite{song2020score} is introduced. Diffusion models have become a prominent framework in generative modeling, achieving notable success across various domains such as image synthesis~\cite{dhariwal2021diffusion}, text generation~\cite{li2022diffusion}, and molecular modeling~\cite{hoogeboom2022equivariant}. Despite their successes, diffusion models often encounter challenges like high computational demands, prolonged generation times, and sub-optimal probability paths~\cite{song2020denoising,lu2022dpm}.

\textbf{Flow matching models}, proposed as an efficient, simulation-free method to train Continuous Normalizing Flows (CNFs), eliminate the need for explicit knowledge of the marginal vector field~\cite{lipman2022flow}. Despite these advancements, conventional solvers still require extensive function evaluations~\cite{dao2023flow}. To address these challenges, several techniques have been developed to improve performance. Rectified Flows (RFs)~\cite{liu2022flow} employ straight-line simulations of probability paths, simplifying training and accelerating sampling by avoiding complex ODE solvers while preventing the collapse of word embeddings. Conditional Flow Matching (CFM)\cite{tong2023conditional} learns the vector field conditioned on individual data points and incorporates Optimal Transport to ensure optimal probability paths, enhancing training efficiency and reducing inference time. While flow-based methods have shown promising results in the design of biological molecules, including protein structures~\cite{yim2024improved}, their application to protein sequence design remains largely unexplored. To bridge this gap, we introduce the first FM-based generative model specifically tailored for protein sequence design.

\section{Background}
\label{sec:background}

\subsection{Problem formulation}
Typically, a protein can be represented as a sequence of amino acids $x=[x_1, x_2, \dots, x_L]$ of length $L$, where each amino acid $x_i$ is selected from a vocabulary $\mathcal{V}$ that includes the 20 standard amino acids. Deep generative models for \textit{de novo} protein sequence design are defined to learn the data distribution $p(x)$ and to sample novel and plausible protein sequences $x$ from this distribution.

\subsection{Preliminaries}

Inspired by score-based generation models~\cite{song2020score}, we can define the mapping between samples from the data distribution $x_0 \sim \pi_0$ and the prior distribution $x_1 \sim \pi_1$ as an ordinary differential equation
\begin{equation} 
d\phi_t = v_\theta(\phi_t, t)dt
\label{eqn1}
\end{equation}
where $\theta$ means the vector field $v:[0,1] \times R^d \rightarrow R^d$ can be learned with a neural network. This vector field defines a unique time-dependent flow $\phi:[0,1] \times R^d \rightarrow R^d$.

The marginals are Gaussian conditioned on some random variable z, which are assigned with concrete meanings later.
\begin{equation} 
p_t(x_t) = \mathbb{E}_{z \sim \mathcal{N}(\mu_t, \sigma_t)}p_t(x|z) = \mathcal{N}(x|\mu_t(z), \sigma_t^2(z))
\label{eqn2}
\end{equation}

One of the simplest flow $\phi$ that can generate this probability path $p_t$ is
\begin{equation} 
\phi_{t,z}(x)=\mu_t(z)+\sigma_t(z)x
\label{eqn3}
\end{equation}

Since the vector field $u_t$ generates the flow, we can learn the flow that matches $p_t$ by using gradient descent in regression against the target vector field, which is called \textbf{Flow Matching}~\cite{lipman2022flow}. That is, We can regress $u_t$ with the \textit{Flow Matching Objective}
\begin{equation} 
\mathcal{L}_{\text{FM}}=\mathbb{E}_{t,p_t(x)}\|v_\theta(x,t)-u_t(x)\|_2^2
\label{eqn4}
\end{equation}

The unique conditional vector field can be written as
\begin{equation} 
u_t(x|z)=\frac{\sigma_t'(z)}{\sigma_t'(z)}(x-\mu_t(z))+\mu_t'(z)
\label{eqn5}
\end{equation}

One can construct the marginal vector field $u_t(x)$ using the conditional vector field $u_t(x|z)$
\begin{equation} 
u_t(x)=\mathbb{E}_{q(z)}\frac{u_t(x|z)p_t(x|z)}{p_t(x)}
\label{eqn6}
\end{equation}

Since $\mathcal{L}_{\text{FM}}$ in intractable due to the marginalization shown in Equation~\ref{eqn6}, people introduce an equivalent tractable \textit{Conitional Flow Matching (CFM) Objective}
\begin{equation} 
\mathcal{L}_{\text{CFM}}=\mathbb{E}_{t,q(z),p_t(x|z)}\|v_\theta(x,t)-u_t(x|z)\|_2^2
\label{eqn7}
\end{equation}

To reduce the complexity of the flow and accelerate the sampling process, \textbf{Rectified Flows (RFs)} \cite{liu2022flow} define the probability paths as straight lines $\mu_t(z)=tx_1+(1-t)x_0$, which further simplify the method into the form
\begin{equation} 
q(z) = \pi_0(x_0)\pi_1(x_1)
\label{eqn8}
\end{equation}
\begin{equation} 
p_t(x|z) = tx_1+(1-t)x_0
\label{eqn9}
\end{equation}
\begin{equation} 
u_t(x|z) = x_1 - x_0
\label{eqn10}
\end{equation}
where $z$ is defined as a tuple of random variables $(x_0, x_1)$, representing samples from data and noise distributions.

Although rectified flows reduce the transport cost, there are still curves and crossings in the probability paths. By using the \textbf{Reflow} technique, where the corresponding pairs $(z_0, z_1)$ with $z_0 \sim \mu_0(z)$ and $z_1 \sim \mu_1(z)$ from the learned rectified flows are sampled and replaced $(x_0, x_1)$ to run new rectified flows, the probability paths can be made nearly straight with further reduced transport cost. In this case, good results can be sampled with even a single ODE solving step.

\section{Methods}
\label{sec:methods}
As depicted in Figure~\ref{fig1}, we design protein sequences with rectified flow-based methods in the redesigned ESM-2 latent space with compression and smoothing. We further accelerate the generation with reflow. Detailed parameter and training settings can be found in Appendix A and algorithms can be found in Appendix B.

\subsection{Design with pLM Latent Space}
To leverage the semantically meaningful latent space of pLM~\cite{hu2024flow}, which is inherently more suitable for continuous FM-based methods, we utilize the pre-trained pLM, ESM-2~\cite{lin2022language}, as the encoder and fine-tune a corresponding decoder. ESM-2 is one of the most advanced pLMs, trained on the large-scale standard protein sequence database Uniref50~\cite{suzek2015uniref}. The encoder maps a protein $x=[x_1, x_2, \dots, x_L] \in \mathcal{V}^L$ into a continuous embedding $h=[h_1, h_2, \dots, h_L] \in \mathcal{R}^{L\times D}$, while the decoder reconstructs the sequence as $x'=[x'_1, x'_2, \dots, x'_L] \in \mathcal{V}^L$. To adapt the pre-trained language model to specific tasks and improve reconstruction accuracy, the decoder is trained independently using a cross-entropy loss between the tokenized $x$ and $x'$. To investigate the impact of pLM model quality, we experiment with two variants of ESM-2 with different parameter scales (8M and 35M). To standardize the length $L$ of all protein sequences, we add padding tokens to the ends of shorter proteins, which will implicitly encode the length information in the latent embedding.

\subsection{Latent Space Redesign}

The ESM-2 embeddings have unnecessarily large dimensions and the massive activation problem~\cite{valeriani2024geometry}, necessitating the redesign of a smoother and more compact latent space. To enhance space efficiency, we introduce a dimensional compressor-decompressor pair positioned between the encoder and decoder. This mechanism compresses the original continuous embedding $h\in\mathcal{R}^{L\times D}$ to $h_c\in \mathcal{R}^{L\times D/c}$ with a compression ratio $c$ and subsequently reconstructs it back to $h'\in\mathcal{R}^{L\times D}$. To adjust the data scale and smooth the data, we add preprocessing operations before the compressor. Since our FM model operates on a standard Gaussian distribution, we apply z-score normalization to scale the output embeddings of ESM-2. The massive activation problem introduces outliers with excessively small variances, which affects the proper convergence. To fix this, we first truncate the z-score normalized embeddings using a saturation function and employ a min-max normalization to smooth the truncated latent space. The corresponding reverse operations are added after the decompressor. With the other parts frozen, the compressor-decompressor pair is trained using an MSE reconstruction objective between $h$ and $h'$.

\subsection{Fast Generation with Flow Matching}

\subsubsection{1-Rectified Flow Generation}

To address the long generation time and suboptimal probability path issues inherent in diffusion-based methods, we adapted the Rectified Flow-based~\cite{liu2022flow} method for protein sequence design. Due to its straight and optimal probability paths, our model can theoretically generate good results with fewer ODE-solving steps.

During training, we randomly sample the sequence $x$, standard Gaussian noise $\epsilon$ and time step $t$ within the time range $[0,1]$. The sequence will be encoded and compressed to $h_c$.
\begin{equation} 
q(z) = \pi_0(h_c)\pi_1(\epsilon)
\label{eqn11}
\end{equation}
Then we calculate $p_t(x|z)$ and the vector field of Rectified Flow $u_t(x|z)$ with
\begin{equation} 
p_t(x|z) = t\epsilon+(1-t)h_c, t \sim \text{Uniform}(0,1)
\label{eqn12}
\end{equation}
\begin{equation} 
u_t(x|z) = \epsilon - h_c
\label{eqn13}
\end{equation}
Finally the model predicts the vector field $v_\theta(h_c,t)$ and trained with the loss function in Equation~\ref{eqn7}.

During inference, we begin by sampling a random standard Gaussian noise $\epsilon$ and iteratively solve the ODE within the learned vector field using an ODE solver. The resulting embedding $h'_c$ is then decompressed using the decompressor and subsequently decoded back into a protein sequence. For 1-Rectified Flow, we employ the dopri5 solver and for the reflow, we use the Euler solver. The optimal number of ODE-solving steps is determined based on specific tasks, with the step range set between $[1,100]$. Although the length of protein sequences is implicitly captured in the latent embedding through the trailing padding tokens, we additionally sample the sequence length from the empirical distribution observed in the validation dataset and apply attention masks to ensure the adequate distribution of generated sequence lengths.

\subsubsection{One-step Generation with Reflow}
To further accelerate the generation, we employ the reflow technique to construct 2-Rectified Flow, which theoretically enables high-quality one-step generation~\cite{liu2022flow}. We sample from the start and end point of the ODE solver for the inference of trained 1-Rectified Flow models. The obtained paired noises and data $(z_0, z_1)$ will replace $(\epsilon, h_c)$ in Equation~\ref{eqn11}-~\ref{eqn13} for finetuning. 

\begin{figure}[t]
\centering
\includegraphics[width=0.81\columnwidth]{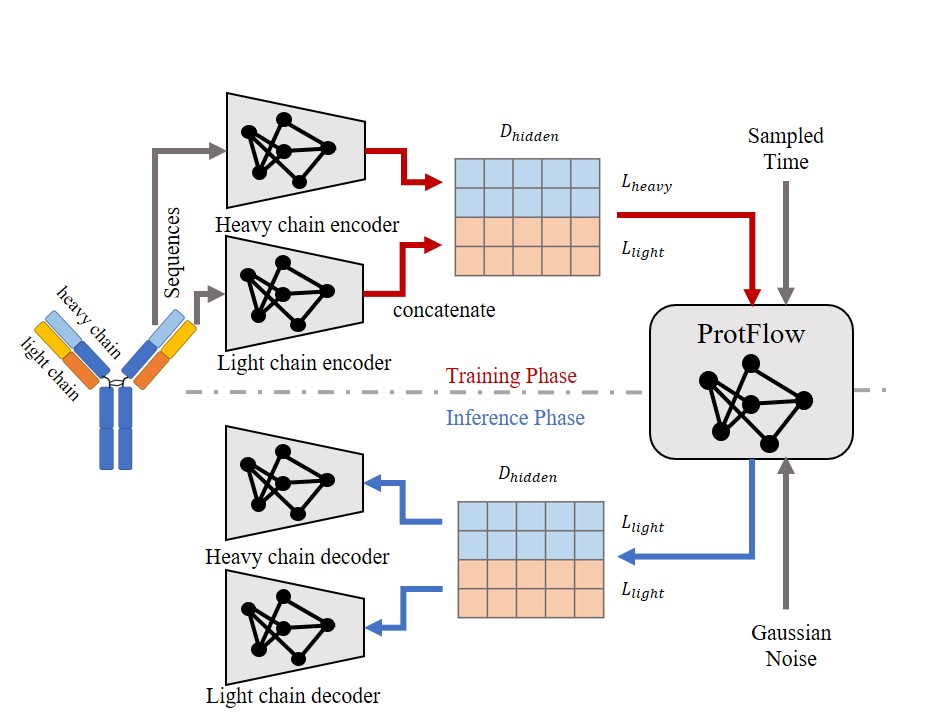} 
\caption{Joint Design of Antibodies. Heavy chains and light chains are utilized to finetune ESM-2 decoders respectively. The embeddings are concatenated before fed into ProtFlow. $D_{hidden}$ is the ESM-2 hidden dimension; $L_{heavy}$ and $L_{light}$ are the maximum lengths of the heavy and light chains.}
\label{fig2}
\end{figure}

\subsection{Multichain Joint Design Pipeline}

For multichain proteins, such as antibodies, significant interdependence often exists among individual chains. Designing each chain independently may compromise the structural integrity of the overall protein. Therefore, we developed a pipeline for the joint design of multichain proteins. Taking antibodies as an example, as shown in Figure~\ref{fig2}, we fine-tune two separate ESM-2 models to encode and decode the heavy chain and the light chain, and concatenate the embeddings to be processed simultaneously by ProtFlow as a whole. During the inference phase, the generated embeddings will be split and independently converted back to protein sequences.

\subsection{Model Architecture}

For the decoder, we employ the same architecture as the ESM-2 LMHead, which consists of two linear layers, an activation layer, and a layer norm.

The compressor consists of two dimension normalization layers, two transformer layers with a pooling layer between them, a down-projection block with a layer norm and a linear layer, and an additional activation layer to smooth the continuous embeddings. The decompressor consists of an unpooling layer, two transformer layers with an up-projection block with a layer norm and a linear layer between them, and corresponding denormalization layers.

The flow-matching holder is constructed as a 12-layer Transformer model. Time information is integrated into the model via a linear projection, and added before each Transformer block. Inspired by the U-Vit architecture~\cite{bao2023all}, and following insights from diffusion-based studies~\cite{meshchaninov2024diffusion}, we utilize long skip connections to enhance the performance and facilitate more effective learning dynamics.

\section{Experiments}
\label{sec:experiments}

In this section, we first determine the optimal compression ratio $c$, and evaluate the model in the general peptide and long-chain protein generation task. Following the impressive results in general tasks, we apply ProtFlow to design a practical functional protein, antimicrobial peptides. Finally, we evaluate the model on a more challenging task, the antibody design, which contains multiple sequences and more complex structures. Please refer to Appendix C for detailed experiment settings.

\subsection{Optimal Compression Ratio}

To ensure the generalizability, we first determine the optimal compression ratio $c$ on general design tasks. We evaluate a range of compression ratios $[1, 2, 4, 8, 16, 32]$ by assessing the reconstruction accuracy and FPD. The results, summarized in Table~\ref{tab:table1}, demonstrate that compression does not significantly impact reconstruction accuracy. As the embedding dimensionality is progressively reduced up to a ratio of 16, the FPD values improve in both cases, indicating an enhanced capability for distribution learning. Beyond this threshold, additional compression may introduce potential trade-offs between efficiency and representation fidelity.

\begin{table}[h!]
\centering
\caption{Reconstruction accuracy and FPD of each Compression Ratio (c) on the general peptide and long-chain protein tasks.}
\resizebox{\linewidth}{!}{
\renewcommand\arraystretch{0.5}
\begin{tabular}{c|cc|cc}
\hline
\multirow{2}{*}{Ratio (c)} & \multicolumn{2}{c|}{Uniprot peptides} & \multicolumn{2}{c}{SwissProt} \\ \cline{2-5} 
                       & Acc(\%)$\uparrow$        & FPD$\downarrow$        & Acc(\%)$\uparrow$         & FPD$\downarrow$         \\
\midrule
1 & 99.93 & 0.48 & 99.98 & 2.09 \\
2 & 99.96 & 0.44 & 99.88 & 1.59 \\
4 & 99.97 & 0.42 & 99.89 & 1.35 \\
8 & 99.96  & 0.41 & 99.87 & 1.23 \\
16 & 99.96 & \textbf{0.36} & 99.91 & \textbf{1.13} \\
32 & 99.98 & 0.43 & 99.91 & 1.26 \\            
\bottomrule

\end{tabular}
}
\label{tab:table1}
\end{table}

\subsection{General Peptide and Protein Design}

This section involves the general protein design cases, including short-chain proteins, i.e. peptides, and long-chain proteins. In these general design cases, we develop ProtFlow and analyze the impacts of each technique of our methods. Experiments demonstrate its versatility and outstanding performance, laying a solid foundation for the subsequent application of ProtFlow in functional proteins.

\begin{table*}[h!]
\centering
\caption{Performance comparison between ProtFlow and baseline models on UniProt Peptides and SwissProt datasets.}
\label{tab:table2}
\resizebox{\linewidth}{!}{%
\renewcommand\arraystretch{0.8}
\begin{tabular}{lccccc|ccc|c}
\toprule
\multirow{2}{*}{} & \multirow{2}{*}{Model} & \multirow{2}{*}{pLDDT ($\uparrow$)} & \multirow{2}{*}{ESM-2 pppl ($\downarrow$)} & \multirow{2}{*}{scPerplexity ($\downarrow$)} & \multirow{2}{*}{TM-score ($\uparrow$)} & \multirow{2}{*}{FPD ($\downarrow$)} & \multirow{2}{*}{MMD ($\downarrow$)} & \multirow{2}{*}{OT ($\downarrow$)} & \multirow{2}{*}{NFE ($\downarrow$)}\\
                  &                        &                     &                           &                             &                       &                     &                      &                   &                   \\
\cmidrule(r){1-10}
\multirow{11}{*}{\rotatebox[origin=c]{90}{\textbf{UniProt Peptides}}} 
& Dataset & 71.35 & 12.58 & 12.31 & 0.57 & 0.23 & 0.000 & 1.74 & nan\\
& Random sequences & 62.89 & 21.71 & 20.40 & 0.44 & 3.66 & 0.084 & 5.75 & nan\\
\cmidrule(r){2-10}
& ProteinGAN & 67.84 & 15.14 & 13.89 & 0.54 & 1.51 & 0.034 & 2.78 & nan\\
& EvoDiff-OADM & 64.99 & 14.26 & 13.61 & 0.50 & 0.60 & 0.014 & 2.29 & 1000\\
& EvoDiff-OADM w/ Transformer & 63.78 & 17.76 & 14.75 & 0.52 & 1.24 & 0.051 & 2.72 & 1000\\
& DiMA & 70.95 & 13.72 & 14.12 & 0.54 & 0.87 & 0.021 & 2.57 & 100\\
& Dirichlet FM & 67.19 & 14.13 & 13.59 & 0.53 & 0.50 & 0.014 & 2.10 & 50\\

\cmidrule(r){2-10}
& ProtFlow w/ ESM2 8M & 71.14 & 12.89 & 13.05 &  0.56 & 0.53 & 0.010 & 2.07 & 25 \\
& ProtFlow w/ ESM-2 35M &  70.54 & 12.96 & 12.83 & 0.56 & 0.46 & 0.009 & 1.99 & 25\\
& ProtFlow w/ ESM-2 35M \& compression& 71.93 & 12.29 & 12.14 &  0.59 & \textbf{0.36} & \textbf{0.004} & \textbf{1.86} & 25 \\
& ProtFlow reflow w/ ESM-2 35M \& compression& \textbf{72.86} & \textbf{12.16} & \textbf{11.79} & \textbf{0.60} & 0.45 & 0.006 & 1.97 & \textbf{1}\\

\midrule

\multirow{7}{*}{\rotatebox[origin=c]{90}{\textbf{SwissProt}}} 
& Dataset          & 79.36 & 9.88 & 5.21 & 0.80 & 0.27 & 0.002 & 1.37 & nan\\
& Random sequences & 26.31 & 21.17 & 14.84 & 0.33 & 3.07 & 0.205 & 3.88 & nan\\
\cmidrule(r){2-10}
& proteinGAN       & 32.77 & 16.12 & 12.36 & 0.24 & 3.08 & 0.176 & 4.02 & nan\\
& EvoDiff-OADM     & 39.36 & 15.97 & 10.76 &  0.2 & 2.17 & 0.112 & 3.50 & 1000\\
& DiMA             & 77.56 & 11.46 & 6.54 & 0.42 & 1.58 & 0.104 & 2.25 & 100\\
& Dirichlet FM     & 53.33 & 12.88 & 9.35 & 0.37 & 1.38 & 0.073 & 2.56 & 50\\
\cmidrule(r){2-10}
& ProtFlow & \textbf{79.65} & \textbf{8.77} & \textbf{4.24} & \textbf{0.47} & \textbf{1.13} & \textbf{0.059} & \textbf{2.35} & \textbf{25}\\

\bottomrule
\end{tabular}
}
\end{table*}

\paragraph{Data}
For general peptides, we collected peptides with lengths of 2 to 50 from UniProt~\cite{uniprot2019uniprot}. After removing sequences with uncommon amino acids, we ended up with 2.6M peptides. For general long proteins, we use \textbf{SwissProt}, a high-quality, manually annotated subset of UniProt. After filtering out sequences shorter than 128 and trimming sequences longer than 254, and removing sequences with uncommon amino acids, we get a total number of 470k proteins.

\paragraph{Baseline}
We include the Generative Adversarial Network (GAN) method ProteinGAN~\cite{repecka2021expanding}, the discrete diffusion method EvoDiff-OADM with ByteNet-CNN backbone and Transformer backbone~\cite{alamdari2023protein}, the continuous diffusion method DiMA~\cite{meshchaninov2024diffusion}, and the discrete FM methods Dirichlet Flow Matching~\cite{stark2024dirichlet}. All baseline models are set with the same or similar number of parameters and use a standard character-level tokenization scheme.

\paragraph{Metrics}
We follow the metrics settings of ~\cite{meshchaninov2024diffusion}. To evaluate sequence quality, we use \textbf{ESM-2 pppl}. To evaluate structural foldability and quality, we use OmegaFold~\cite{wu2022high} \textbf{pLDDT} and ESM-IF \textbf{scPerplexity}. To evaluate naturalness, we use \textbf{TM-Score} against SwissProt. To reflect the ability to capture the data distribution, we use \textbf{Frechet ProtT5 Distance (FPD)}, \textbf{Maximum mean discrepancy (MMD)} and \textbf{1-Wasserstein optimal
transport (OT)}. For the generation speed, we use the \textbf{Number of Function Evaluations (NFE)}.

\paragraph{Results}
The results are presented in Table~\ref{tab:table2}. We highlight our main findings as follows:

\paragraph{(1) ProtFlow generates high-quality sequences with leading distributions.} 
We investigate the quality of the sequences ProtFlow generates leveraging ESM-2 Perplexity and ESM-IF scPerplexity. ESM-2 Perplexity reflects how well a given sequence aligns with the patterns ESM-2 has learned from its training data, while ESM-IF scPerplexity estimates the quality and reliability on the structural level. ProtFlow achieves the best perplexity results better than the dataset, revealing that it is capable of generating protein sequences with high reliability and good-alignment with natural protein patterns. FPD, MMD, and OT measure the distributional similarity between the generated sequences and the training data. ProtFlow outperforms all baselines. This indicates that ProtFlow is a good data distribution learner. Notably, when the model gains better distribution values, its perplexities may exhibit slight degradation. This might indicate the trade-off between the computing reliability and the residue-level diversity of protein sequences.

\paragraph{(2) ProtFlow generates sequences with foldable and natural predicted structures.}
The predicted local distance difference test (pLDDT) score is the most authoritative measurement of the structural plausibility of protein sequences. Typically, a protein with a pLDDT score greater than 70 is considered to have high structural confidence. After predicting the structure with OmegaFold, we compare the structural similarity against known structures in SwissProt PDB with TM-Score. ProtFlow generates protein sequences with the highest pLDDT score and TM-Score, which are comparable to or even better than the datasets for the general protein design tasks. This demonstrates that ProtFlow generates sequences with high foldability and structural naturalness. 

\paragraph{(3) Flow matching leads fast and optimal generation.}
One of the most significant advantages of FM is its fast generation speed.Unlike diffusion models, which require hundreds or even thousands of steps, Dirichlet FM achieves comparable results with only 50 steps, making it twice as efficient as DiMA. Leveraging the nearly linear probability path of 1-RF, ProtFlow achieves excellent results with just 25 steps, achieving a fourfold speed improvement over DiMA. Reflow further straightens the probability path, enabling even single-step generation using the simplest Euler ODE solver. Although fine-tuning based on imprecise 1-RF probability path sampling may slightly compromise distributional performance, reflow still outperforms all other methods. Moreover, by learning the optimal probability path, FM ensures excellent generation quality while maintaining fast generation speed. The most direct manifestation of this is the superiority of FM-based methods in distributional metrics. Notably, both Dirichlet FM and ProtFlow, especially the model using the same ESM2-8M with DiMA, excel beyond the diffusion models, EvoDiff and DiMA, which operate in discrete and continuous spaces, respectively. Specifically, while DiMA achieves impressive scores in the initial metrics, it performs less favorably in terms of distribution metrics compared to both flow-based methods. This consistency underscores the superior performance of FM methods, by learning and modeling optimal probability paths, excel particularly in capturing accurate data distributions. 

\paragraph{(4) Latent space redesign improves FM effects.}
As a method operating on the continuous latent space built by the pLM ESM-2, ProtFlow outperforms Dirichlet FM, which directly operates on discrete sequences; Another latent method, DiMA, also achieves better scores on foldability, reliability and naturalness than discrete methods, EvoDiff and Dirichlet FM. This advantage partly reflects the benefits of working within a more condensed and semantically rich latent space. In Table~\ref{tab:table1}, as the dimensionality is progressively compressed, the model's ability to learn the data distribution improves until a 16-fold compression, where excessive compression likely leads to the loss of possible probability path information, reducing the model's performance. Notably, performing FM in the compressed latent space allows the model to outperform the datasets. This indicates that latent space compression not only makes the model more compact in terms of spatial representation, but also enables more streamlined information, which further facilitates the learning of optimal probability paths through FM.

\subsection{Antimicrobial Peptide Design}

Antimicrobial peptides (AMPs) are a class of functional proteins with antimicrobial properties that play a crucial role in the innate immune response of mammals against invasive bacterial infections. Currently, researchers in the biochemical field are focused on developing novel and robust antimicrobial peptides, a process that generative models can assist in and accelerate. The experiments demonstrated that ProtFlow is capable of creating high-quality, novel, diverse AMPs, highlighting its significant potential to design proteins with other desired attributes for practical applications.

\paragraph{Data}
Following AMP-Diffusion~\cite{chen2024amp}, we collected AMPs from databases dbAMP~\cite{jhong2019dbamp}, AMP Scanner~\cite{veltri2018deep}, and DRAMP~\cite{kang2019dramp}. After filtering out sequences with length greater than 40 and containing uncommon amino acids, expanding by HydrAMP~\cite{szymczak2023discovering} and removing duplicate sequences, 195k AMPs are left for training.


\paragraph{Baseline}
We utilize baselines from AMP-Diffusion, including two advanced conditional VAE methods HydrAMP~\cite{szymczak2023discovering} and PepCVAE~\cite{das2018pepcvae}, the GAN method AMPGAN~\cite{van2021ampgan}, and the diffusion method AMP-Diffusion. 

\paragraph{Metrics}
We also follow the metrics settings of AMP-Diffusion. To evaluate the sequence quality of the generated proteins, we use \textbf{ESM-2 pppl}. To evaluate the diversity and similarity of the generated proteins, \textbf{Shannon Entropy} and 6-mers \textbf{Jaccard Similarity} (JS-6) are utilized. Finally we use an external classifier of HydrAMP to calculate the proportion of generated peptides that can be classied as AMPs, \textbf{$P_{amp}$} and anti-\textit{E. coli}, \textbf{$P_{mic}$}.

\begin{table}[h!]
\centering
\caption{Performance on AMP design of ProtFlow and baseline methods on the AMP dataset. $\dagger$: benchmarked results are quoted from [Chen et al., 2024].} 
\resizebox{\linewidth}{!}{
\renewcommand\arraystretch{1.5}

\begin{tabular}{lccccc} 
\toprule
{Model} & {Perplexity($\downarrow$)} & {Entropy($\uparrow$)} & {JS-6($\uparrow$)} & $P_\text{amp}$($\uparrow$) & $P_\text{mic}$($\uparrow$) \\

\midrule
Train/Real AMPs$\dagger$       & 16.23 $\pm$ 3.80 & 3.23 $\pm$ 0.40  & - & - & - \\
\midrule
AMPGAN$\dagger$  & 17.70 $\pm$ 4.08 & 2.90 $\pm$ 0.61  & 0.016 & 0.54 & 0.32 \\
PepCVAE$\dagger$  & 19.82 $\pm$ 3.83 & 3.12 $\pm$ 0.36  & 0.020 & 0.41 & 0.20 \\
HydrAMP$\dagger$  & 17.27 $\pm$ 6.02 & 2.82 $\pm$ 0.48  & 0.014 & 0.77 & 0.49 \\
AMP-Diffusion$\dagger$  & 12.84 $\pm$ 4.53 & 3.17 $\pm$ 0.56 & 0.028 & 0.81 & 0.50 \\
\midrule
ProtFlow  & \textbf{12.24 $\pm$ 3.26} & \textbf{3.19 $\pm$ 0.55}  & \textbf{0.080} & \textbf{0.84} & \textbf{0.63}  \\
\bottomrule
\end{tabular}
}
\label{tab:table3}
\end{table}

\paragraph{Results}
Table~\ref{tab:table3} delineates the performance of ProtFlow in comparison to other relevant models. We find that ProtFlow outperforms other AMP-specific methods. The lowest ESM-2 perplexity demonstrates that ProtFlow can generate highly reliable peptides. Shannon entropy serves as a measure of the sequence’s uncertainty or randomness, reflecting its information content and complexity. ProtFlow gains the highest entropy, indicating that it can generate the most diverse peptide sequences. Meanwhile, ProtFlow receives a remarkably higher 6-mer JS score, indicating its ability to learn complex sequence motifs relevant to AMP design. 

For the antimicrobial property classification test, we take a threshold for {$P_{amp}$} as 0.8, which can be a good likelihood of being antimicrobial. Since the MIC property is a subset of the antimicrobial property, and the peptides experimentally measured MIC to train the HydrAMP classifier is limited \cite{szymczak2023discovering}, a threshold for {$P_{mic}$} set as 0.5 is enough to show the peptide activity potential against \textit{E. coli}. ProtFlow generates the notably highest proportion of peptides exceeding both thresholds, meaning ProtFlow can generate peptides not only being classified as AMPs, but also exhibiting effective activity on specific pathogenic bacteria.

\subsection{Antibody Design}

Antibodies are large Y-shaped proteins composed of heavy and light chains with important immune functions. The design of novel antibodies has long been a critical focus in bioengineering and medicine. However, given the complexity of antibodies, designing them requires the simultaneous generation of both the heavy and light chains. Leveraging our multichain joint design pipeline, ProtFlow is able to operate on this more complicated task. Through antibody design, we have demonstrated that ProtFlow is capable of designing large, complex proteins and can collaboratively design multiple related proteins. This further expands the potential of ProtFlow for practical application. We follow the benchmark settings of L-WJS~\cite{mahajan2023exploiting}.

\paragraph{Data}
 We collected 1.4M pairs of antibodies (containing heavy and light chains) from the Observed Antibody Space (OAS) database \cite{olsen2022observed}. Sequences were clustered at 95$\%$ sequence identity with 80$\%$ coverage using MMseqs2 \cite{steinegger2017mmseqs2}. Sequences with uncommon amino acids, heavy chain longer than 149 and light chain longer than 148 are filtered out. 

\paragraph{Baseline}
We use baselines including the language models GPT3.5, IgLM~\cite{shuai2021generative} and ESM-2~\cite{lin2023evolutionary}, an energy-based method DEEN~\cite{saremi2018deep}, and the diffusion methods dWJS~\cite{frey2023protein} and L-WJS~\cite{mahajan2023exploiting}. 

\paragraph{Metrics}
To reflect the learning quality of the training data distribution on the physicochemical property scale, we use the Wasserstein Distance $W_{property}$ between the sampled sequences and the reference distribution on 15 biological properties. To evaluate the novelty, we utilize the proportion of unique sequences (Uniqueness). To evaluate sequence diversity, we use the average edit distance of sampled sequences from the reference distribution, $E_{dist}$, and the average edit distance within the generated sequences, IntDiv. 

\paragraph{Results}
Table~\ref{tab:table4} exhibits the performance of ProtFlow in comparison to other relevant models. Notably, ESM-2 achieves the highest $E_{dist}$ and IntDiv. However, as mentioned in dWJS \cite{frey2023protein}, ESM-2 does not generate antibody-like sequences, making these leading scores meaningless. ProtFlow receives the lowest $W_{property}$, representing its promising learning ability of the biophysical property distributions of the paired OAS. Most of the models show good Uniqueness generation. ProtFlow gains the second best IntDiv and $E_{dist}$ score values, which are still competitive. It might reflect the trade-off between the distribution learning and novelty together with diversity.

\begin{table}[h!]
\centering
\caption { Performance on antibody design of ProtFlow and baseline methods on the OAS dataset. The \textbf{best} and \underline{second best} results are marked. * means not desirable results with high diversity without matching
the reference distribution. $\dagger$: benchmarked results are quoted from [Mahajan et al., 2023] and [Frey et al., 2023].}
\resizebox{\linewidth}{!}{
\renewcommand\arraystretch{1.0}
\tabcolsep=0.015cm 

\begin{tabular}{lcccc}
\toprule
{Model} & $W_{\text{property}} \downarrow$ & {Uniqueness $\uparrow$} & $E_{\text{dist}} (\hat{z}, x) \uparrow$ & {IntDiv $\uparrow$} \\

\midrule
dWJS $\dagger$ & 0.065  & 0.97 & \textbf{62.7} & \textbf{65.1}  \\
L-WJS $\dagger$ & \underline{0.053}  & \textbf{1.0}   & 56.6    & 54.1  \\
DEEN $\dagger$ & 0.087  & \underline{0.99}   & 50.9    & 42.7  \\
GPT3.5 $\dagger$ & 0.140  & 0.66 & 55.4    & 46.1  \\
IgLM $\dagger$ & 0.087  & \textbf{1.0}  & 48.6    & 34.6  \\
ESM2 $\dagger$ & 0.150   & \textbf{1.0}  & 70.99*  & 77.56* \\
\midrule
ProtFlow & \textbf{0.045}  & \textbf{1.0}   & \underline{58.6}    & \underline{58.98}  \\
\bottomrule
\end{tabular}
}
\label{tab:table4}
\end{table}

\section{Conclusions}
\label{sec:conclusions}

In this work, we introduced ProtFlow, the first flow matching-based model for \textit{de novo} protein sequence generation. Our approach leverages the straight and optimal probability path of rectified flow to achieve fast and high-quality sequence generation. By operating on a compressed and smoothed latent space derived from protein language models (pLMs), we enhanced the efficiency and accuracy of our method. Furthermore, the Reflow technique enabled high-quality one-step generation, significantly reducing computational costs. Additionally, we developed a multichain joint design pipeline and applied ProtFlow to various protein design tasks. ProtFlow consistently achieves state-of-the-art performance across all evaluated tasks with minimal generation steps, demonstrating its strong capability in rapidly designing high-quality protein sequences. For future work, exploration of scaling data, conditional generation strategies and multi-modality such as protein structure data can be introduced to extend the potential applications of ProtFlow to a much broader range.
\clearpage
\bibliographystyle{named}

\begin{thebibliography}{}

\bibitem[\protect\citeauthoryear{Alamdari \bgroup \em et al.\egroup }{2023}]{alamdari2023protein}
Sarah Alamdari, Nitya Thakkar, Rianne van~den Berg, Alex~Xijie Lu, Nicolo Fusi, Ava~Pardis Amini, and Kevin~K Yang.
\newblock Protein generation with evolutionary diffusion: sequence is all you need.
\newblock {\em bioRxiv}, pages 2023--09, 2023.

\bibitem[\protect\citeauthoryear{Bao \bgroup \em et al.\egroup }{2023}]{bao2023all}
Fan Bao, Shen Nie, Kaiwen Xue, Yue Cao, Chongxuan Li, Hang Su, and Jun Zhu.
\newblock All are worth words: A vit backbone for diffusion models.
\newblock In {\em Proceedings of the IEEE/CVF conference on computer vision and pattern recognition}, pages 22669--22679, 2023.

\bibitem[\protect\citeauthoryear{Brown}{2020}]{brown2020language}
Tom~B Brown.
\newblock Language models are few-shot learners.
\newblock {\em arXiv preprint ArXiv:2005.14165}, 2020.

\bibitem[\protect\citeauthoryear{Chen \bgroup \em et al.\egroup }{2024}]{chen2024amp}
Tianlai Chen, Pranay Vure, Rishab Pulugurta, and Pranam Chatterjee.
\newblock Amp-diffusion: Integrating latent diffusion with protein language models for antimicrobial peptide generation.
\newblock {\em bioRxiv}, pages 2024--03, 2024.

\bibitem[\protect\citeauthoryear{Consortium}{2019}]{uniprot2019uniprot}
UniProt Consortium.
\newblock Uniprot: a worldwide hub of protein knowledge.
\newblock {\em Nucleic acids research}, 47(D1):D506--D515, 2019.

\bibitem[\protect\citeauthoryear{Dao \bgroup \em et al.\egroup }{2023}]{dao2023flow}
Quan Dao, Hao Phung, Binh Nguyen, and Anh Tran.
\newblock Flow matching in latent space.
\newblock {\em arXiv preprint arXiv:2307.08698}, 2023.

\bibitem[\protect\citeauthoryear{Das \bgroup \em et al.\egroup }{2018}]{das2018pepcvae}
Payel Das, Kahini Wadhawan, Oscar Chang, Tom Sercu, Cicero dos Santos, Matthew Riemer, Vijil Chenthamarakshan, Inkit Padhi, and Aleksandra Mojsilovic.
\newblock Pepcvae: Semi-supervised targeted design of antimicrobial peptide molecules.
\newblock {\em arXiv preprint arXiv:1810.07743}, 2018.

\bibitem[\protect\citeauthoryear{Dhariwal and Nichol}{2021}]{dhariwal2021diffusion}
Prafulla Dhariwal and Alexander Nichol.
\newblock Diffusion models beat gans on image synthesis.
\newblock {\em Advances in neural information processing systems}, 34:8780--8794, 2021.

\bibitem[\protect\citeauthoryear{Eijkelboom \bgroup \em et al.\egroup }{2024}]{eijkelboom2024variational}
Floor Eijkelboom, Grigory Bartosh, Christian~Andersson Naesseth, Max Welling, and Jan-Willem van~de Meent.
\newblock Variational flow matching for graph generation.
\newblock {\em arXiv preprint arXiv:2406.04843}, 2024.

\bibitem[\protect\citeauthoryear{Frey \bgroup \em et al.\egroup }{2023}]{frey2023protein}
Nathan~C Frey, Daniel Berenberg, Karina Zadorozhny, Joseph Kleinhenz, Julien Lafrance-Vanasse, Isidro Hotzel, Yan Wu, Stephen Ra, Richard Bonneau, Kyunghyun Cho, et~al.
\newblock Protein discovery with discrete walk-jump sampling.
\newblock {\em arXiv preprint arXiv:2306.12360}, 2023.

\bibitem[\protect\citeauthoryear{Ho \bgroup \em et al.\egroup }{2020}]{ho2020denoising}
Jonathan Ho, Ajay Jain, and Pieter Abbeel.
\newblock Denoising diffusion probabilistic models.
\newblock {\em Advances in neural information processing systems}, 33:6840--6851, 2020.

\bibitem[\protect\citeauthoryear{Hoogeboom \bgroup \em et al.\egroup }{2022}]{hoogeboom2022equivariant}
Emiel Hoogeboom, V{\i}ctor~Garcia Satorras, Cl{\'e}ment Vignac, and Max Welling.
\newblock Equivariant diffusion for molecule generation in 3d.
\newblock In {\em International conference on machine learning}, pages 8867--8887. PMLR, 2022.

\bibitem[\protect\citeauthoryear{Hu \bgroup \em et al.\egroup }{2024}]{hu2024flow}
Vincent Hu, Di~Wu, Yuki Asano, Pascal Mettes, Basura Fernando, Bj{\"o}rn Ommer, and Cees Snoek.
\newblock Flow matching for conditional text generation in a few sampling steps.
\newblock In {\em Proceedings of the 18th Conference of the European Chapter of the Association for Computational Linguistics (Volume 2: Short Papers)}, pages 380--392, 2024.

\bibitem[\protect\citeauthoryear{Huang \bgroup \em et al.\egroup }{2016}]{huang2016coming}
Po-Ssu Huang, Scott~E Boyken, and David Baker.
\newblock The coming of age of de novo protein design.
\newblock {\em Nature}, 537(7620):320--327, 2016.

\bibitem[\protect\citeauthoryear{Jhong \bgroup \em et al.\egroup }{2019}]{jhong2019dbamp}
Jhih-Hua Jhong, Yu-Hsiang Chi, Wen-Chi Li, Tsai-Hsuan Lin, Kai-Yao Huang, and Tzong-Yi Lee.
\newblock dbamp: an integrated resource for exploring antimicrobial peptides with functional activities and physicochemical properties on transcriptome and proteome data.
\newblock {\em Nucleic acids research}, 47(D1):D285--D297, 2019.

\bibitem[\protect\citeauthoryear{Kang \bgroup \em et al.\egroup }{2019}]{kang2019dramp}
Xinyue Kang, Fanyi Dong, Cheng Shi, Shicai Liu, Jian Sun, Jiaxin Chen, Haiqi Li, Hanmei Xu, Xingzhen Lao, and Heng Zheng.
\newblock Dramp 2.0, an updated data repository of antimicrobial peptides.
\newblock {\em Scientific data}, 6(1):148, 2019.

\bibitem[\protect\citeauthoryear{Li \bgroup \em et al.\egroup }{2022}]{li2022diffusion}
Xiang Li, John Thickstun, Ishaan Gulrajani, Percy~S Liang, and Tatsunori~B Hashimoto.
\newblock Diffusion-lm improves controllable text generation.
\newblock {\em Advances in Neural Information Processing Systems}, 35:4328--4343, 2022.

\bibitem[\protect\citeauthoryear{Lin \bgroup \em et al.\egroup }{2022}]{lin2022language}
Zeming Lin, Halil Akin, Roshan Rao, Brian Hie, Zhongkai Zhu, Wenting Lu, Allan dos Santos~Costa, Maryam Fazel-Zarandi, Tom Sercu, Sal Candido, et~al.
\newblock Language models of protein sequences at the scale of evolution enable accurate structure prediction.
\newblock {\em BioRxiv}, 2022:500902, 2022.

\bibitem[\protect\citeauthoryear{Lin \bgroup \em et al.\egroup }{2023}]{lin2023evolutionary}
Zeming Lin, Halil Akin, Roshan Rao, Brian Hie, Zhongkai Zhu, Wenting Lu, Nikita Smetanin, Robert Verkuil, Ori Kabeli, Yaniv Shmueli, et~al.
\newblock Evolutionary-scale prediction of atomic-level protein structure with a language model.
\newblock {\em Science}, 379(6637):1123--1130, 2023.

\bibitem[\protect\citeauthoryear{Lipman \bgroup \em et al.\egroup }{2022}]{lipman2022flow}
Yaron Lipman, Ricky~TQ Chen, Heli Ben-Hamu, Maximilian Nickel, and Matt Le.
\newblock Flow matching for generative modeling.
\newblock {\em arXiv preprint arXiv:2210.02747}, 2022.

\bibitem[\protect\citeauthoryear{Liu \bgroup \em et al.\egroup }{2022}]{liu2022flow}
Xingchao Liu, Chengyue Gong, and Qiang Liu.
\newblock Flow straight and fast: Learning to generate and transfer data with rectified flow.
\newblock {\em arXiv preprint arXiv:2209.03003}, 2022.

\bibitem[\protect\citeauthoryear{Lu \bgroup \em et al.\egroup }{2022}]{lu2022dpm}
Cheng Lu, Yuhao Zhou, Fan Bao, Jianfei Chen, Chongxuan Li, and Jun Zhu.
\newblock Dpm-solver: A fast ode solver for diffusion probabilistic model sampling in around 10 steps.
\newblock {\em Advances in Neural Information Processing Systems}, 35:5775--5787, 2022.

\bibitem[\protect\citeauthoryear{Lu \bgroup \em et al.\egroup }{2024}]{lu2024tokenized}
Amy~X Lu, Wilson Yan, Kevin~K Yang, Vladimir Gligorijevic, Kyunghyun Cho, Pieter Abbeel, Richard Bonneau, and Nathan Frey.
\newblock Tokenized and continuous embedding compressions of protein sequence and structure.
\newblock {\em bioRxiv}, pages 2024--08, 2024.

\bibitem[\protect\citeauthoryear{Mahajan \bgroup \em et al.\egroup }{2023}]{mahajan2023exploiting}
Sai~Pooja Mahajan, Nathan~C Frey, Daniel Berenberg, Joseph Kleinhenz, Richard Bonneau, Vladimir Gligorijevic, Andrew Watkins, and Saeed Saremi.
\newblock Exploiting language models for protein discovery with latent walk-jump sampling.
\newblock {\em Unknown}, 2023.

\bibitem[\protect\citeauthoryear{Meshchaninov \bgroup \em et al.\egroup }{2024}]{meshchaninov2024diffusion}
Viacheslav Meshchaninov, Pavel Strashnov, Andrey Shevtsov, Fedor Nikolaev, Nikita Ivanisenko, Olga Kardymon, and Dmitry Vetrov.
\newblock Diffusion on language model embeddings for protein sequence generation.
\newblock {\em arXiv preprint arXiv:2403.03726}, 2024.

\bibitem[\protect\citeauthoryear{Olsen \bgroup \em et al.\egroup }{2022}]{olsen2022observed}
Tobias~H Olsen, Fergus Boyles, and Charlotte~M Deane.
\newblock Observed antibody space: A diverse database of cleaned, annotated, and translated unpaired and paired antibody sequences.
\newblock {\em Protein Science}, 31(1):141--146, 2022.

\bibitem[\protect\citeauthoryear{Repecka \bgroup \em et al.\egroup }{2021}]{repecka2021expanding}
Donatas Repecka, Vykintas Jauniskis, et~al.
\newblock Expanding functional protein sequence spaces using generative adversarial networks.
\newblock {\em Nature Machine Intelligence}, 3(4):324--333, 2021.

\bibitem[\protect\citeauthoryear{Rombach \bgroup \em et al.\egroup }{2022}]{rombach2022high}
Robin Rombach, Andreas Blattmann, Dominik Lorenz, Patrick Esser, and Bj{\"o}rn Ommer.
\newblock High-resolution image synthesis with latent diffusion models.
\newblock {\em Unknown}, pages 10684--10695, 2022.

\bibitem[\protect\citeauthoryear{Ruffolo and Madani}{2024}]{ruffolo2024designing}
Jeffrey~A Ruffolo and Ali Madani.
\newblock Designing proteins with language models.
\newblock {\em nature biotechnology}, 42(2):200--202, 2024.

\bibitem[\protect\citeauthoryear{Saremi \bgroup \em et al.\egroup }{2018}]{saremi2018deep}
Saeed Saremi, Arash Mehrjou, Bernhard Sch{\"o}lkopf, and Aapo Hyv{\"a}rinen.
\newblock Deep energy estimator networks.
\newblock {\em arXiv preprint arXiv:1805.08306}, 2018.

\bibitem[\protect\citeauthoryear{Shuai \bgroup \em et al.\egroup }{2021}]{shuai2021generative}
Richard~W Shuai, Jeffrey~A Ruffolo, and Jeffrey~J Gray.
\newblock Generative language modeling for antibody design.
\newblock {\em BioRxiv}, pages 2021--12, 2021.

\bibitem[\protect\citeauthoryear{Song \bgroup \em et al.\egroup }{2020a}]{song2020denoising}
Jiaming Song, Chenlin Meng, and Stefano Ermon.
\newblock Denoising diffusion implicit models.
\newblock {\em arXiv preprint arXiv:2010.02502}, 2020.

\bibitem[\protect\citeauthoryear{Song \bgroup \em et al.\egroup }{2020b}]{song2020score}
Yang Song, Jascha Sohl-Dickstein, Diederik~P Kingma, Abhishek Kumar, Stefano Ermon, and Ben Poole.
\newblock Score-based generative modeling through stochastic differential equations.
\newblock {\em arXiv preprint arXiv:2011.13456}, 2020.

\bibitem[\protect\citeauthoryear{Stark \bgroup \em et al.\egroup }{2024}]{stark2024dirichlet}
Hannes Stark, Bowen Jing, Chenyu Wang, Gabriele Corso, Bonnie Berger, Regina Barzilay, and Tommi Jaakkola.
\newblock Dirichlet flow matching with applications to dna sequence design.
\newblock {\em arXiv preprint arXiv:2402.05841}, 2024.

\bibitem[\protect\citeauthoryear{Steinegger and S{\"o}ding}{2017}]{steinegger2017mmseqs2}
Martin Steinegger and Johannes S{\"o}ding.
\newblock Mmseqs2 enables sensitive protein sequence searching for the analysis of massive data sets.
\newblock {\em Nature biotechnology}, 35(11):1026--1028, 2017.

\bibitem[\protect\citeauthoryear{Suzek \bgroup \em et al.\egroup }{2015}]{suzek2015uniref}
Baris~E Suzek, Yuqi Wang, Hongzhan Huang, Peter~B McGarvey, Cathy~H Wu, and UniProt Consortium.
\newblock Uniref clusters: a comprehensive and scalable alternative for improving sequence similarity searches.
\newblock {\em Bioinformatics}, 31(6):926--932, 2015.

\bibitem[\protect\citeauthoryear{Szymczak \bgroup \em et al.\egroup }{2023}]{szymczak2023discovering}
Paulina Szymczak, Marcin Mo{\.z}ejko, Tomasz Grzegorzek, Rados{\l}aw Jurczak, Marta Bauer, Damian Neubauer, Karol Sikora, Micha{\l} Michalski, Jacek Sroka, Piotr Setny, et~al.
\newblock Discovering highly potent antimicrobial peptides with deep generative model hydramp.
\newblock {\em nature communications}, 14(1):1453, 2023.

\bibitem[\protect\citeauthoryear{Tong \bgroup \em et al.\egroup }{2023}]{tong2023conditional}
Alexander Tong, Nikolay Malkin, Guillaume Huguet, Yanlei Zhang, Jarrid Rector-Brooks, Kilian Fatras, Guy Wolf, and Yoshua Bengio.
\newblock Conditional flow matching: Simulation-free dynamic optimal transport.
\newblock {\em arXiv preprint arXiv:2302.00482}, 2(3), 2023.

\bibitem[\protect\citeauthoryear{Valeriani \bgroup \em et al.\egroup }{2024}]{valeriani2024geometry}
Lucrezia Valeriani, Diego Doimo, Francesca Cuturello, Alessandro Laio, Alessio Ansuini, and Alberto Cazzaniga.
\newblock The geometry of hidden representations of large transformer models.
\newblock {\em Advances in Neural Information Processing Systems}, 36, 2024.

\bibitem[\protect\citeauthoryear{Van~Oort \bgroup \em et al.\egroup }{2021}]{van2021ampgan}
Colin~M Van~Oort, Jonathon~B Ferrell, Jacob~M Remington, Safwan Wshah, and Jianing Li.
\newblock Ampgan v2: machine learning-guided design of antimicrobial peptides.
\newblock {\em Journal of chemical information and modeling}, 61(5):2198--2207, 2021.

\bibitem[\protect\citeauthoryear{Veltri \bgroup \em et al.\egroup }{2018}]{veltri2018deep}
Daniel Veltri, Uday Kamath, and Amarda Shehu.
\newblock Deep learning improves antimicrobial peptide recognition.
\newblock {\em Bioinformatics}, 34(16):2740--2747, 2018.

\bibitem[\protect\citeauthoryear{Wang \bgroup \em et al.\egroup }{2024}]{wang2024diffusion}
Xinyou Wang, Zaixiang Zheng, Fei Ye, Dongyu Xue, Shujian Huang, and Quanquan Gu.
\newblock Diffusion language models are versatile protein learners.
\newblock {\em arXiv preprint arXiv:2402.18567}, 2024.

\bibitem[\protect\citeauthoryear{Wu \bgroup \em et al.\egroup }{2022}]{wu2022high}
Ruidong Wu, Fan Ding, Rui Wang, Rui Shen, Xiwen Zhang, Shitong Luo, Chenpeng Su, Zuofan Wu, Qi~Xie, Bonnie Berger, et~al.
\newblock High-resolution de novo structure prediction from primary sequence.
\newblock {\em BioRxiv}, pages 2022--07, 2022.

\bibitem[\protect\citeauthoryear{Yeh \bgroup \em et al.\egroup }{2023}]{yeh2023novo}
Andy Hsien-Wei Yeh, Christoffer Norn, Yakov Kipnis, Doug Tischer, Samuel~J Pellock, Declan Evans, Pengchen Ma, Gyu~Rie Lee, Jason~Z Zhang, Ivan Anishchenko, et~al.
\newblock De novo design of luciferases using deep learning.
\newblock {\em Nature}, 614(7949):774--780, 2023.

\bibitem[\protect\citeauthoryear{Yim \bgroup \em et al.\egroup }{2024}]{yim2024improved}
Jason Yim, Andrew Campbell, Emile Mathieu, Andrew~YK Foong, Michael Gastegger, Jos{\'e} Jim{\'e}nez-Luna, Sarah Lewis, Victor~Garcia Satorras, Bastiaan~S Veeling, Frank No{\'e}, et~al.
\newblock Improved motif-scaffolding with se (3) flow matching.
\newblock {\em ArXiv}, 2024.

\bibitem[\protect\citeauthoryear{Zhu \bgroup \em et al.\egroup }{2024a}]{zhu2024generative}
Yiheng Zhu, Zitai Kong, Jialu Wu, Weize Liu, Yuqiang Han, Mingze Yin, Hongxia Xu, Chang-Yu Hsieh, and Tingjun Hou.
\newblock Generative ai for controllable protein sequence design: A survey.
\newblock {\em arXiv preprint arXiv:2402.10516}, 2024.

\bibitem[\protect\citeauthoryear{Zhu \bgroup \em et al.\egroup }{2024b}]{zhu2024bridgeif}
Yiheng Zhu, Jialu Wu, Qiuyi Li, Jiahuan Yan, Mingze Yin, Wei Wu, Mingyang Li, Jieping Ye, Zheng Wang, and Jian Wu.
\newblock Bridge-{IF}: Learning inverse protein folding with markov bridges.
\newblock In {\em The Thirty-eighth Annual Conference on Neural Information Processing Systems}, 2024.

\end{thebibliography}

\clearpage
\appendix

\section{Models \& Settings}

\subsection{Model Architecture}

For the compressor and decompressor, we modified based on the settings of Hourglass Transformer~\cite{lu2024tokenized}. We utilize the hidden size of 480 corresponding to the embedding of ESM-2 35M, the depth of 4, 8 attention heads, 64 dimensional heads and no hidden dropout rate. The pooling layer uses mean pooling and the activation layer uses tanh function. The linear layers make projections between hidden size of 480 and the compressed hidden size of $480/c$. The two additional sets of norm and denorm layers are dimensional z-score and min-max normalizations.

The backbone of the flow matching holder module is a 12-layer BERT model with 16 attention heads. We use the GELU activation function, with an intermediate layer size of 3072 and an attention dropout rate of 0.1. Depending on the ESM-2 variant employed, the hidden size is set to 320 when using ESM-2 8M and 480 when using ESM-2 35M, with a hidden dropout rate of 0.1. To match the compressed hidden size of the compressor and decompressor, we add two additional linear layer to make projections between hidden size of 480 and the compressed hidden size of $480/c$. The noised protein sequence embeddings are added with positional encodings before being fed into the flow matching holder module. Timesteps are projected to match the size of the protein sequence embeddings using a sinusoidal embedding block, and these are added to the input of each Transformer block after a linear projection. We maintain the advanced long skip connections from ~\cite{meshchaninov2024diffusion}, which involve adding the linear projections of earlier block inputs to those of later blocks.

\subsection{Experimental Settings}

For fine-tuning the ESM-2 decoder, the model is trained for 1 epoch with 1000 steps, and validation is performed every 50 steps. The learning rate is set to 0.00005. We use a batch size of 512 and the AdamW optimizer, with beta values of (0.9, 0.98) and a weight decay of 0.001.

For training of the compressor and decompressor, all experiments converge within 30 epoches. We perform validation using reconstrction accuracy on the validation set and save checkpoints every 5000 iterations. A cosine learning rate scheduler with two cycle limit is employed, featuring a minimum learning rate of 8e-5 and 10000 linear warmup steps. The batch size is set to 16 for both training and validation. The AdamW optimizer is configured with beta values of (0.9, 0.999).

For training of the flow matching holder, all experiments converge within 1,000,000 iterations. We perform validation using FPD on the validation set and save checkpoints every 10,000 iterations. A cosine learning rate scheduler with a single cycle limit is employed, featuring a minimum learning rate of 0.0002 and 5000 linear warmup steps. The batch size is set to 64 for peptides and AMPs training and validation, and set to 16 for general proteins and antibodies. The AdamW optimizer is configured with beta values of (0.9, 0.98), the weight decay of 0.01, the epsilon of 0.000001, and the gradient clipping norm of 1. We trained on one NVIDIA GeForce RTX 3090 GPU. Training models on the general peptide and AMP task for 1,000,000 training steps takes around 42 hours while the general protein and antibody task needs 120 hours.

For inference, we test the Euler ODE solver with $N \in \{1, 5, 10, 25\}$ steps for reflow and the Dormand Prince 45 ODE solver with $N \in \{1, 5, 10, 25, 50, 75, 100\}$ steps for the other tasks. The optimal solver and sampling steps are different for different tasks. Normally the Dormand Prince 45 ODE solver with a sampling step of 25 can give relatively good results with robustness. For reflow task, the model can support 1-step generation.

\section{Algorithms}

\subsection{Training procedure}

\begin{algorithm}
\caption{Training ProtFlow with 1-RF}
\begin{algorithmic}[1]
\STATE \textbf{Input:} data distribution $\pi_0$, noise distribution $\epsilon$, initial neural network $v_\theta$, encoder $Emb$, compressor $C$, training step $T$.
\STATE \textbf{Output:} trained neural network $v_\theta$.
\FOR{$1,2,\dots, i$ in $T$}
\STATE    $x_0 \sim \pi_0$, $x_1 \sim \epsilon$
\STATE    $x_0 \leftarrow \text{padding}(x_0)$
\STATE    $x_0 \leftarrow Emb(x_0)$
\STATE    $x_0 \leftarrow C(x_0)$
\STATE    $t \sim \text{Uniform}(0,1)$
\STATE    $x_t \leftarrow tx_1 + (1-t)x_0$
\STATE    $u_t \leftarrow x_1 - x_0$
\STATE    $\mathcal{L}_{\text{CFM}}(\theta) \leftarrow \|v_\theta(x_t,t)-u_t\|_2^2$
\STATE    $ \theta \leftarrow \text{Update}(\nabla\mathcal{L}_{\text{CFM}}(\theta), \theta)$
\ENDFOR
\STATE Return $v_\theta$.
\end{algorithmic}
\end{algorithm}

\subsection{Inference procedure}

\begin{algorithm}
\caption{ProtFlow Sampling}
\begin{algorithmic}[1]
\STATE \textbf{Input:} data distribution $\pi_0$, noise distribution $\epsilon$, neural network $v_\theta$, decompressor $DeC$, decoder $D$, sampling step $N$.
\STATE \textbf{Output:} generated samples $x_0$.
\STATE    $x_1 \sim \epsilon$
\STATE    $x_t \leftarrow x_1$
\FOR{$1,2,\dots, i$ in $N$}
\STATE    $t \leftarrow 1/i$
\STATE    $x_t \leftarrow \text{ODE\_Solver}(v_\theta(x_t, t), x_t, N)$
\ENDFOR
\STATE    $x_0 \leftarrow x_t$
\STATE    $x_0 \leftarrow DeC(x_0)$
\STATE    $mask \sim \text{Length\_Sampler}(pi_0)$
\STATE    $x_0 \leftarrow D(x_0, mask)$
\STATE Return $x_0$.
\end{algorithmic}
\end{algorithm}

\subsection{ReFlow Training procedure}
\begin{algorithm}
\caption{Training ProtFlow with 1-RF}
\begin{algorithmic}[1]
\STATE \textbf{Input:} noise distribution $\epsilon$, trained 1-RF neural network $v_{1-RF}$, encoder $Emb$, compressor $C$, training step $T$, sampling step $M$, sampling step $N$.
\STATE \textbf{Output:} trained reflow neural network $v_\theta$.
\STATE    $\pi \sim \{\}$
\FOR{$1,2,\dots, i$ in $M$}
\STATE    $x_1 \sim \epsilon$
\STATE    $x_t \leftarrow x_1$
\FOR{$1,2,\dots, i$ in $N$}
\STATE    $t \leftarrow 1/i$
\STATE    $x_t \leftarrow \text{ODE\_Solver}(v_\theta(x_t, t), x_t, N)$
\ENDFOR
\STATE    $z_0 \sim x_0$, $z_1 \sim x_1$
\STATE    $\pi.append((z_0,z_1))$
\ENDFOR
\FOR{$1,2,\dots, i$ in $T$}
\STATE    $z_0,z_1 \sim \pi$
\STATE    $t \sim \text{Uniform}(0,1)$
\STATE    $z_t \leftarrow tz_1 + (1-t)z_0$
\STATE    $u_t \leftarrow z_1 - z_0$
\STATE    $\mathcal{L}_{\text{CFM}}(\theta) \leftarrow \|v_\theta(z_t,t)-u_t\|_2^2$
\STATE    $ \theta \leftarrow \text{Update}(\nabla\mathcal{L}_{\text{CFM}}(\theta), \theta)$
\ENDFOR
\STATE Return $v_\theta$.
\end{algorithmic}
\end{algorithm}

\section{Experiment Details}

\subsection{General Protein Design}




\subsubsection{Baselines}

\textbf{ProteinGAN}~\cite{repecka2021expanding} represents a pioneering application of generative models in \textit{de novo} protein sequence design. It is a variant of the generative adversarial network (GAN), where both the discriminator and generator are based on ResNet-based convolutional neural networks (CNNs), further enhanced with a self-attention layer. While ProteinGAN was originally developed for Multiple Sequence Alignment (MSA) tasks, which focus on specific protein families, it has also demonstrated competitive performance in our general peptide design task.

\textbf{EvoDiff-OADM}~\cite{alamdari2023protein} is the first foundation diffusion model for protein design trained on evolutionary-scale protein sequence data. EvoDiff grounds in two famous discrete diffusion frameworks: D3PM and OADM, while the OADM-based model is reported to perform better in the original paper. EvoDiff-OADM operates in an order-agnostic autoregressive manner, gradually converting single amino acids to or from the mask token during the forward or reverse processes. The model is based on the ByteNet CNN architecture, and we utilize a configuration with 38 million parameters. Furthermore, we evaluate its performance after replacing the backbone with the Transformer-based ESM-2 35M architecture.

\textbf{DiMA}~\cite{meshchaninov2024diffusion} is one of the earliest diffusion model operating on continuous space by embedding discrete protein sequences with the protein language model ESM-2 8M and surpasses lots of traditional mainstream generative models. It is built on a 12-layer transformer model and introduces multiple novel techniques to improve the performance, including long skip connections, self-conditioning and tanh noise schedule. 

\textbf{Dirichlet Flow Matching}~\cite{stark2024dirichlet} is a pioneering approach in discrete flow matching, framing generative modeling as a transportation problem on the probability simplex and using the Dirichlet distribution as the probability distribution for the generative path. It has demonstrated strong performance in DNA sequence design tasks. We adapted Dirichlet Flow Matching from the DNA sequence design into the amino acid sequence generation of protein, retaining the original CNN architecture and model hyperparameter settings.



\subsubsection{Metrics}

\textbf{Predicted Local Distance Difference Test (pLDDT)} is a metric used to assess the accuracy of protein structure predictions. It reflects the foldability and  structural plausibility of a protein sequence by predicting the confidence score for each residue position in the protein structure. We utilize OmegaFold~\cite{wu2022high} to predict the pLDDT of each residue of the sequence and average them as the pLDDT of that sequence.

\textbf{ESM-2 pseudoperplexity (ESM-2 pppl)} measures how well a given protein sequence aligns with the patterns the assessing model has learned from the training data of ESM-2, which is a large scale standard protein database UniRef50~\cite{suzek2015uniref}. We calculate ESM-2 pppl with ESM-2 35M~\cite{lin2023evolutionary} by masking each amino acid of the protein sequence and predicting it considering all the other amino acids in the sequence, and calculating the value with the equation
\begin{equation*}
    \text{ESM-2 pseudoperplexity} = \exp{( -\frac{1}{|x|} \sum_{i=1}^{|x|} \log p(x_i \mid x_{j \neq i}, \theta_{\text{ESM-2}}) )}
\end{equation*}
where $|x|$ means the length of sequence x and $x_{j \neq i}$ means sequence x without the $i^th$ amino acids.

\textbf{Self-consistency perplexity (scPerplexity)} provides a measurement of sequence reliability and quality in a structure-based aspect. After predicting the structure of each sequence with OmegaFold~\cite{wu2022high}, we utilize ESM-IF 142M to inverse fold the structure, which predicts a sequence that would naturally fold into that structure. Then we compute the perplexity against the original generated sequence.

\textbf{TM-Score} evaluates the similarity between two protein pairs in structure. Compared to pLDDT which also make evaluations on the structure level, TM-Score focuses more on the global level. For each protein sequence, after predicting the structure with OmegaFold, we use the FoldSeek easy-search tool to seek for the most similar protein in the AlphaFold SwissProt database with the TM-Score function
\begin{equation*}
    \text{TM-score} = \frac{1}{|x_{query}|} \sum_{i=1}^{|x_{query}|} \frac{1}{1 + \left(\frac{d_i}{\sigma|x_{target}|}\right)^2}
\end{equation*}
where $|x_{query}|$ means the number of aligned residues between the query and target proteins, $|x_{target}|$ is the length of the target protein, $\sigma$ is a scaling factor, and $d_i$ is the distance between the $i^{th}$ aligned residue pairs.

\textbf{Fr\'echet ProtT5 Distance (FPD)} is a variant of Fr\'echet distance (FD), which measures the dissimilarity between two samples drawn from multivariate Gaussian distributions. Given two samples $X_1 \sim \mathcal{N}(\mu_1, \Sigma_1)$ and $X_2 \sim \mathcal{N}(\mu_2, \Sigma_2)$, the FID can be calculated as
\begin{equation*}
    \text{Fr\'echet Distance} = \lVert \mu_1 - \mu_2 \rVert^2 + \text{tr}(\Sigma_1 + \Sigma_2 - 2\sqrt{\Sigma_1 \Sigma_2})
\end{equation*}
We calculate the Fr\'echet distance of protein sequence embeddings using protein language model ProtT5, namely, Fr\'echet ProtT5 Distance (FPD).

\textbf{Maximum mean discrepancy (MMD)} is a kernel-based statistical test to determine whether two samples $X=\{x_1,x_2,\dots,x_n\}$ and $Y=\{y_1,y_2,\dots,y_n\}$belong to different distributions. Suppose the kernel is k, then MMD can be calculated as
\begin{equation*}
    \text{MMD} = \frac{1}{n^2} \sum_{i=1}^{n} \sum_{j=1}^{n} \left( k(x_i, x_j) + k(y_i, y_j) - 2k(x_i, y_j) \right)
\end{equation*}
We calculate MMD on the ProtT5 embeddings and use the radial basis function kernel.

\textbf{1-Wasserstein optimal transport (OT)} evaluates the similarity between two batches of sequences. We use pairwise Levenshtein distances as transportation costs, utilize the Earth Mover Distance (EMD) solver with a uniform distribution of the samples to determine optimal sequence pairs, and take the average of the distances between optimal pairs.

\subsection{Antimicrobial Peptide Design}

\subsubsection{Baselines}

\textbf{HydrAMP}~\cite{szymczak2023discovering} is based on a conditional variational autoencoder (cVAE) with an autoencoder and a decoder, and incorporates a pre-trained classifier. It captures the antimicrobial properties of peptides by learning their low-dimensional, continuous representations and decouples these properties from the antimicrobial conditions. The model is designed to generate peptide sequences that meet specific antimicrobial activity criteria. HydrAMP can perform both unconstrained generation and generate antimicrobial analogs based on provided prototype peptides.

\textbf{PepCVAE}~\cite{das2018pepcvae} is a semi-supervised generative model also based on a cVAE and incorporates a pre-trained Antimicrobial Peptide (AMP) classifier. The model learns a rich latent space representation by leveraging a large dataset of unlabeled peptide sequences along with a smaller set of sequences labeled as either antimicrobial or non-antimicrobial. By decoupling the antimicrobial properties from the latent space, PepCVAE is able to generate peptide sequences that meet specific antimicrobial objectives.

\textbf{AMPGAN}~\cite{van2021ampgan} is based on a Bidirectional Conditional Generative Adversarial Network (BiCGAN). In addition to the standard generator and discriminator components found in typical GANs, it introduces an encoder that maps real data into the generator’s latent space. This enables iterative peptide sequence generation, optimization, interpolation, and incremental modifications. The generation process in AMPGAN is controlled by conditioning variables, allowing the discriminator to learn the relationship between relevant antimicrobial features and the generated peptide sequences.

\textbf{AMP-Diffusion}~\cite{chen2024amp} is built on the continuous diffusion model framework, where peptide sequences are mapped into a latent space using the protein language model ESM-2 for the diffusion process. The model architecture incorporates pre-trained ESM-2 8M attention blocks for the denoising process and a multilayer perceptron (MLP) as the output layer. The timestep is embedded with a positional encoding and integrated into protein embeddings with a scaling factor and a bias adjustment. AMP-Diffusion reports a leading performance compared to mainstream AMP design methods and is able to generate AMPs with good predicted activities.

\subsubsection{Metrics}

\textbf{ESM-2 pseudoperplexity} is utilized again to evaluate the sequence quality. The measurement settings keep the same with the ESM-2 pppl of the general peptide design task.

\textbf{Shannon Entropy} is a concept from information theory measuring the uncertainty of information source or randomness of information, reflecting the information content and complexity. In protein sequence design tasks, Shannon entropy can indicate the diversity of single protein sequence. For each peptide sequence, we compute the probability of occurrence of each amino acid $p(x_i), i \in \{0, 1, 2, \dots, 19\}$, and calculate the Shannon entropy as
\begin{equation*}
    H(X) = -\sum_{i=1}^{n} p(x_i) \log_2 p(x_i)
\end{equation*}

\textbf{Jaccard Similarity Coefficient} is a widely used measurement to evaluate the similarity of two sets. We use 6-mers Jaccard similarity coefficient (JS-6) to compare the similarity of generated peptide sets to the training data. We split the training dataset and the generated peptide set with 6-mers, which represents all the sub-sequences with length 6 in the dataset to be processed. Then the Jaccard similarity can be calculated with the equation
\begin{equation*}
    \text{Jaccard}(A, B) = \frac{|A \cap B|}{|A \cup B|}
\end{equation*}
where A represents the 6-mers set of training dataset and B is the 6-mers set of generated batch set.

\textbf{External Classifier} can be an additional validation tool to ascertain the antimicrobial properties of generated peptides. We use the classifier of HydrAMP ~\cite{szymczak2023discovering}, which consists of two distinct networks predicting the probability of a peptide being antimicrobial, and the minimal inhibitory concentration (MIC) against \textit{E. coli}, a significant kind of bacteria serving as a model organism. We utilize the HydrAMP classifiers to score each peptide generated, and calculate the proportion of generated peptides with AMP score greater than 0.8 as \textbf{$P_{amp}$} and MIC score greater than 0.5 as \textbf{$P_{mic}$}.

\subsection{Antibody Design}

\subsubsection{Baselines}

\textbf{dWJS}~\cite{frey2023protein} is a discrete generative model that combines Energy-Based Models (EBMs) with score-based models. The dWJS method is built upon a novel Smoothed Discrete Sampling (SDS) framework. The model trains the EBM by applying maximum likelihood estimation on noisy data and introduces a Walk-Jump sampling-denoising mechanism. Specifically, it utilizes Langevin MCMC to sample from the smoothed noisy data distribution, followed by a denoising step using a separately trained neural network. dWJS is primarily applied in antibody design, where it efficiently generates high-quality and novel samples.

\textbf{L-WJS}~\cite{mahajan2023exploiting} introduces a single-step, score-based diffusion framework for antibody sequence design from higher dimensional embeddings of pretrained language models (pLMs). It also utilize the Walk-Jump sampling-denoising mechanism of dWJS but it sampling on the continuous space of pLMs.

representations of sequences. All modules are trained simultaneously.

\textbf{DEEN}~\cite{saremi2018deep} can be thought of as an energy parameterization of a score-based diffusion model. DEEN is based on MLPs that are used to estimate the energy of the data distribution. This energy function is learned by minimizing a scalable objective function that is constructed using score matching, thereby avoiding the direct computation of the partition function. Moreover, DEEN is not trained with contrastive divergence, so the EBM formulation is completely distinct in terms of parameterization, training, and sampling.

\textbf{IgLM}~\cite{shuai2021generative} is a GPT2-style transformer-based language model trained specifically for antibody design. It was trained on 558M antibody heavy chain and light chain sequences from the Observed Antibody Space (OAS) database. The input sequences to the model are tagged with conditional labels that indicate the type of chain (heavy or light) and the source species, allowing the generated sequences to be controlled based on specific species and chain types.

\textbf{ESM-2}~\cite{lin2023evolutionary} is a large-scale protein language model with 15B parameters. It is built on a Transformer architecture and employs relative position encoding to handle sequences of varying lengths, enhancing the model's ability to generalize across sequences of any length. The model is trained on protein sequence data using the Masked Language Model (MLM) task, where it predicts the identity of randomly masked amino acids. ESM-2 is capable of making highly accurate protein structure predictions based solely on the protein sequence. However, dWJS~\cite{frey2023protein} reports that since ESM-2 is not trained for antibody generation, it generates highly repetitive sequences that are very dissimilar to antibodies. Then its leading metric scores are therefore meaningless.

Additionally, the famous large language model on general tasks, GPT3.5, is also taken into consideration of the baseline.

\subsubsection{Metrics}

\textbf{Properties Wasserstein distance ($W_{property}$)} is the normalized average Wasserstein distance between the property distributions of generated samples and the validation set, reflecting the learning quality of the training data distribution. For each antibody sequence of the heavy chain, we first calculate fifteen biological propeties with the biopython, including the sequence length, molecular weight, aromaticity, instability index, isoelectric point, gravy, charge at pH6, charge at pH7, helix fraction, turn structure fraction, sheet structure fraction, molar extinction coefficient reduced, molar extinction coefficient oxidized, average hydrophilicity, and average surface accessibility. For each property distribution of the generated sequences and validation datasets, we calculate the Wasserstein distance using scipy after minmax normalization. The final Wasserstein distance is the average of the Wasserstein distance of these 15 property distributions.

\textbf{Uniqueness} is a simple indicator evaluating the novelty of the generated sequences. It is the fraction of unique sequences in the generated sequences.

\textbf{Edit distance ($E_{dist}$)}, also known as the Levenshtein distance, is a metric used to measure the similarity between two strings. It represents the minimum number of edit operations required to transform one string into another. The edit distance of sequence $s_1$ and $s_2$ can be calculated use the recursive formula

\begin{equation*}
    d(i, j) = 
\begin{cases} 
0 & \text{if } i = 0 \text{ and } j = 0 \\
j & \text{if } i = 0 \\
i & \text{if } j = 0 \\
d(i-1, j) + 1 & \text{if } s_1[i] \neq s_2[j] \text{ and replace} \\
d(i, j-1) + 1 & \text{if } s_1[i] \neq s_2[j] \text{ and insert} \\
d(i-1, j-1) & \text{if } s_1[i] = s_2[j] \\
\end{cases}
\end{equation*}
where $d(i,j)$ means the minimum number of edit operations required to transform the first $i$ characters of $s_1$ into the first $j$ characters of $s_2$. We utilize the mean of the edit distance of sampled sequences from the validation set, which summarizes the novelty and diversity of samples compared to the validation set.

\textbf{Internal diversity (IntDiv)} is the mean of the edit distance within the generated sequences, reflecting the diversity within the distribution of generated samples. For each sequence, we calculate the edit distance from the rest of the generated sequences and take the average. 

\section{Additional Visualization of Generated Proteins}

\begin{figure}[t]
\centering
\includegraphics[width=0.9\columnwidth]{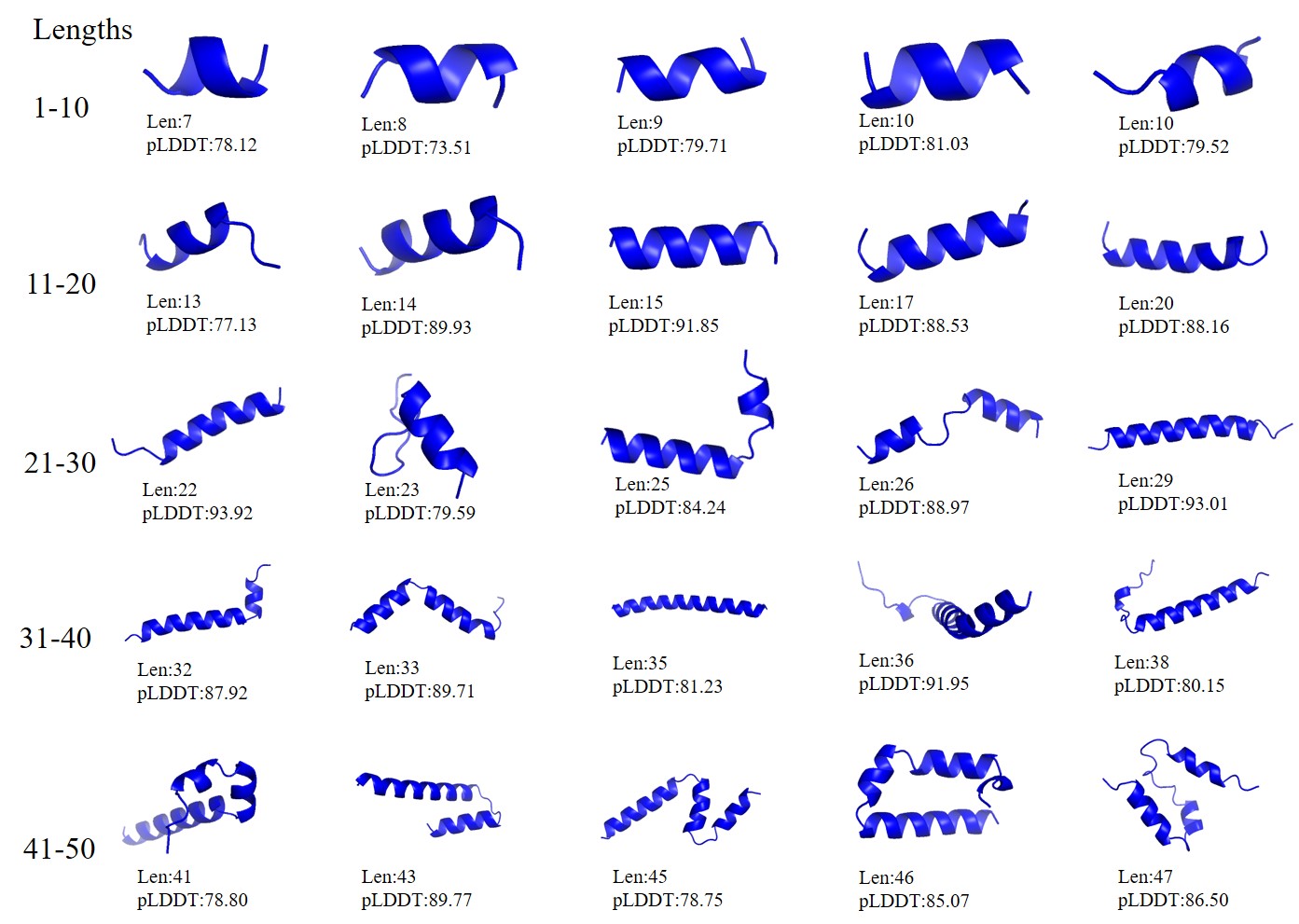} 
\caption{Visualized examples of general peptide design.}
\label{fig3}
\end{figure}

\begin{figure}[t]
\centering
\includegraphics[width=0.9\columnwidth]{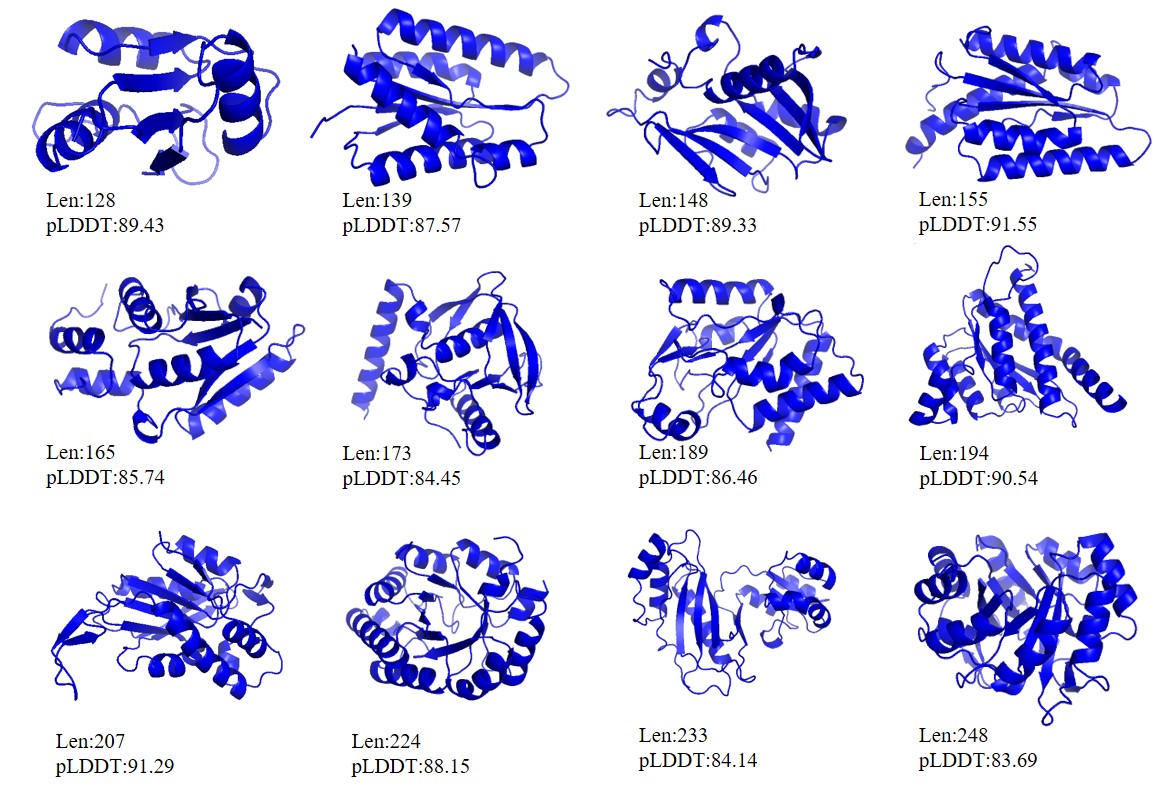} 
\caption{Visualized examples of general long-chain protein design.}
\label{fig4}
\end{figure}

\begin{figure}[t]
\centering
\includegraphics[width=0.9\columnwidth]{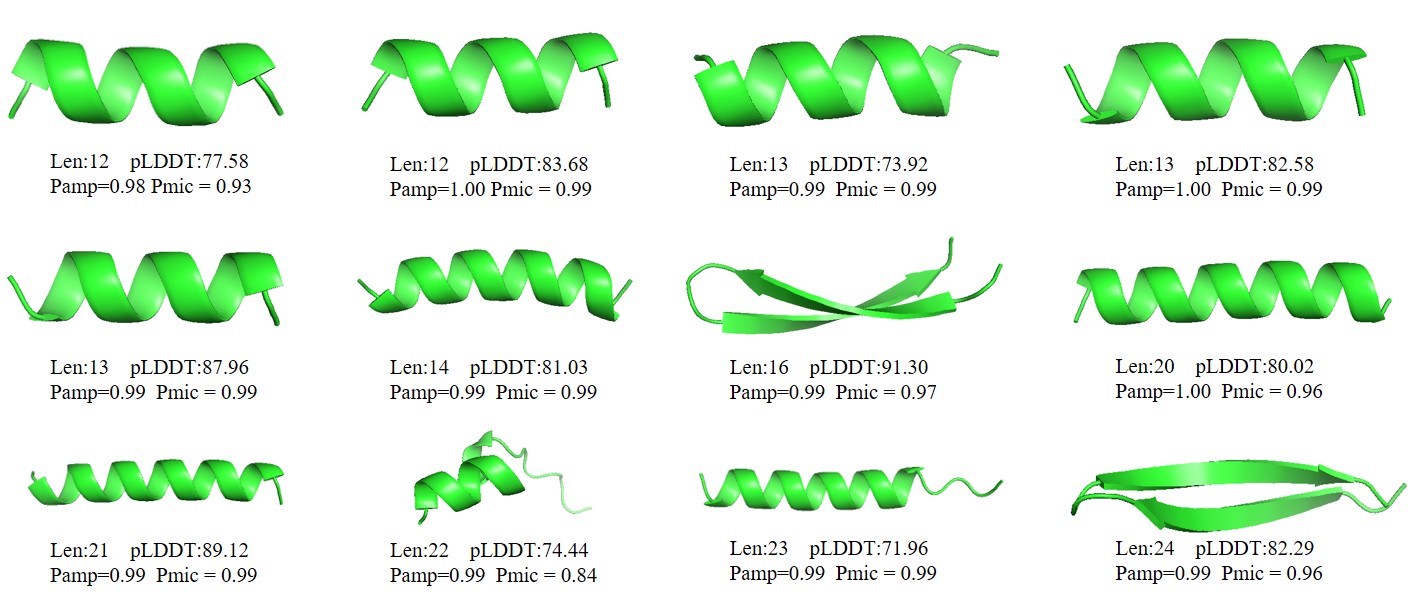} 
\caption{Visualized examples of AMP design.}
\label{fig5}
\end{figure}

\begin{figure}[t]
\centering
\includegraphics[width=0.9\columnwidth]{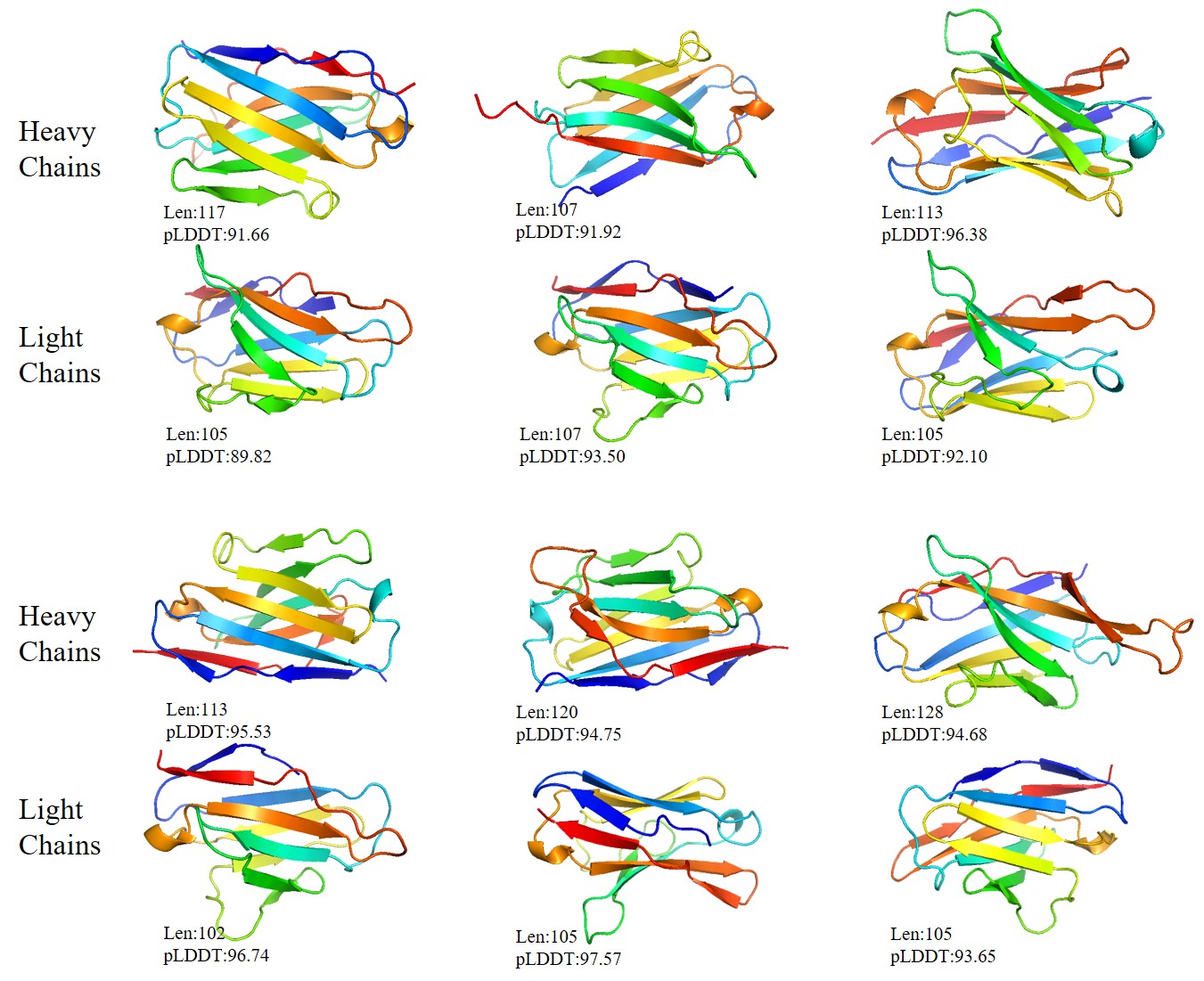} 
\caption{Visualized examples of antibody design.}
\label{fig6}
\end{figure}

\end{document}